# myNER: Contextualized Burmese Named Entity Recognition with Bidirectional LSTM and fastText Embeddings via Joint Training with POS Tagging

Kaung Lwin Thant
VMES
Assumption University
Bangkok, Thailand
Email: g6519696@au.edu

Kwankamol Nongpong
VMES
Assumption University
Bangkok, Thailand
Email: kwankamolnng@au.edu

Ye Kyaw Thu
NECTEC, NSTDA
LU. Laboratory
Pathum Thani, Thailand
Email: yekyaw.thu@nectec.or.th

Thura Aung
School of Engineering
KMITL
Bangkok, Thailand
Email: 66011606@kmitl.ac.th

Khaing Hsu Wai
Graduate School of Engineering Science
Akita University
Akita, Japan
Email: khainghsuwai@gipc.akita-u.ac.jp

Thazin Myint Oo
LU. Laboratory
Yangon, Myanmar
Email: queenofthazin@gmail.com

*Abstract*—Named Entity Recognition (NER) involves identifying and categorizing named entities within textual data. Despite its significance, NER research has often overlooked low-resource languages like Myanmar (Burmese), primarily due to the lack of publicly available annotated datasets. To address this, we introduce *myNER*, a novel word-level NER corpus featuring a 7-tag annotation scheme, enriched with Part-of-Speech (POS) tagging to provide additional syntactic information. Alongside the corpus, we conduct a comprehensive evaluation of NER models, including Conditional Random Fields (CRF), Bidirectional LSTM (BiLSTM)-CRF, and their combinations with fastText embeddings in different settings. Our experiments reveal the effectiveness of contextualized word embeddings and the impact of joint training with POS tagging, demonstrating significant performance improvements across models. The traditional CRF joint-task model with fastText embeddings as a feature achieved the best result, with a 0.9818 accuracy and 0.9811 **weighted F1 score** with 0.7429 **macro F1 score**. BiLSTM-CRF with fine-tuned fastText embeddings gets the best result of 0.9791 accuracy and 0.9776 weighted F1 score with 0.7395 macro F1 score.

## I. Introduction

Named Entity Recognition (NER) is a fundamental task in natural language processing (NLP) that involves identifying and categorizing named entities such as people, organizations, locations, dates, and other domain-specific terms within text. NER plays a crucial role in applications like information extraction, machine translation, question answering, and search engines. While significant advancements have been made in NER systems for high-resource languages, low-resource languages like Myanmar (Burmese) remain underrepresented in NER research. This is primarily due to the lack of annotated datasets and resources necessary for training and evaluating NER models. Addressing these challenges is critical to extending the benefits of NLP technologies to low-resource languages and increasing their accessibility to a broader audience.

Hsu et al. (2022) [1] highlighted the superiority of syllable-based neural models over traditional statistical CRF models, eliminating the need for extensive feature engineering. Using character-level feature engineering with a Convolutional Neural Network (CNN) and a BiLSTM-CRF model trained with the Adam optimizer, they achieved precision, recall, and F1-scores of 95.04%, 94.97%, and 94.39%, respectively. Their work focused on a syllable-level tagged NER dataset with six tag sets: PNAME (Person), LOC (Location), ORG (Organization), RACE, TIME, and NUM (Number).

For Burmese Named Entity Recognition (NER), Conditional Random Fields (CRFs) have be widely adopted [6]. A common approach in NER involves combining CRFs with variants of Long Short-Term Memory (LSTM) networks [7]. Yang [8] systematically reproduced key prior research on sequence labeling, comparing the performance of Softmax and CRF in two inference layers.

In this paper, we introduce a **word-level** NER corpus enriched with Part-of-Speech (POS) tagging to explore the joint behavior and its impact on these tasks. Bidirectional LSTM networks with Conditional Random Fields (CRF) [2] and Softmax as an inference layer have been used for sequence labeling tasks in the Myanmar language for POS Tagging [3] and Sentence Segmentation [4]. Traditional CRF models with and without fastText word embeddings is used for machine learning and BiLSTM models with CRF and Softmax inference layers are used for deep learning models for both single-task and joint-task approaches.

## II. Corpus Development

myNER corpus containing 16,605 sentences, originally sourced from data taken from the myPOS [3] version 3 corpus, is manually annotated NER corpus for our experiments. In our corpus, we used the BIOES (B-Begin, I-Inside, O-Outside, E-End, S-Single) tagging scheme with a total of seven tag categories as shown in Table I.

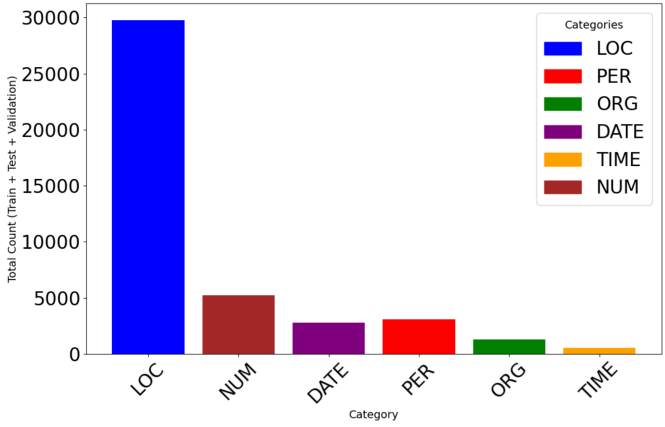

Fig. 1: Tag categories except main O Tag in the dataset

Figure 1 shows the number of tag categories in the dataset that excludes the main tag O. Table II presents the count of each tag and its respective subtags across the train, validation, and test data splits used in the experiments. For the syntactic (POS Tag) information we use myPOS which contains 15 tag sets - abb (Abbreviation), adj (Adjective), adv (Adverb), conj (Conjunction), fw (Foreign word), int (Interjection), n (Noun), num (Number), part (Particle), ppm (Post-positional Marker), pron (Pronoun), punc (Punctuation), sb (Symbol), tn (Text Number), and v (Verb).

Table III illustrates the dataset formatted in CoNLL style with the sentence မန္တလေး ၌ ရန်ကုန် တက္ကသိုလ် လက်အောက်ခံ ဆေးအတတ်သင် ကောလိပ် ရှိ ခဲ့ သည် (In Mandalay, there was a subordinate Medical College of Yangon University). In this example, မန္တလေး (Mandalay) and ရန်ကုန် (Yangon) are locations. However, based on the context, we tagged ရန်ကုန် တက္ကသိုလ် as an ORG entity.

| Word | POS Tag | NER Tag |
|---|---|---|
| မန္တလေး | n | S-LOC |
| ၌ | ppm | O |
| ရန်ကုန် | n | B-ORG |
| တက္ကသိုလ် | n | E-ORG |
| လက်အောက်ခံ | n | O |
| ဆေးအတတ်သင် | n | B-ORG |
| ကောလိပ် | n | E-ORG |
| ရှိ | v | O |
| ခဲ့ | part | O |
| သည် | ppm | O |

TABLE III: myNER sample data with CoNLL format

## III. Methodologies

In this study, we explore the performance of sequence tagging methodologies, machine learning-based traditional CRF, deep learning-based BiLSTM-CRF, and BiLSTM-Softmax models as shown in Figure 2 for word-level NER single-task and joint POS+NER tasks.

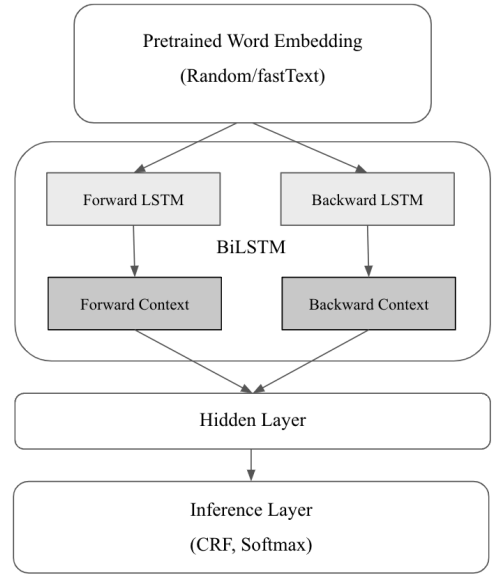

Fig. 2: Word Embeddings + BiLSTM Sequence Tagging

### A. Word Embeddings

fastText [5] from FAIR (Facebook AI Research) is an advanced word embedding technique developed by Facebook AI Research. It extends the traditional Word2Vec model by representing words as bags of character n-grams, which allows it to capture subword information and handle out-of-vocabulary (OOV) words effectively. For each word, fastText generates a dense vector representation by averaging the embeddings of its constituent character n-grams. This approach is particularly useful for morphologically rich languages such as the Myanmar language and improves the representation of rare words. In our study, we use pre-trained fastText embeddings to initialize the word vectors by both fine-tuning and not fine-tuning. We also used random initialization of word embeddings to compare with the pretrained fastText model.

TABLE I: Tag Names, Descriptions, and Examples of myNER Corpus

| Tag | Description | Examples |
|---|---|---|
| DATE | Date | ၁ ရက် ၅ လ ၁၉၉၆ (01-05-1996), ကဆုန်လပြည့် (full moon day of Kasone) |
| TIME | Time | မနက် ၆ နာရီ ၁၀ မိနစ် (06:10 AM) |
| NUM | Numbers | ၁ (1), ၁သန်း (1 Million) |
| PER | Person Names | အန်တိုနီယို (Antonio), ကောင်းလွင်သန့် (Kaung Lwin Thant) |
| ORG | Organizations and Institutions | မိုက်ခရိုဆော့ (Microsoft Inc), ဟားဗတ်တက္ကသိုလ် (Harvard University) |
| LOC | Locations and Geographic Features | အာရှတိုက် (Asia), ထိုင်းနိုင်ငံ (Thailand) |
| O | Outside | Nonspecific entities which do not match the above tags |

TABLE II: Tag Counts for Train, Valid, and Test Datasets with BIES subtags

| Tags | Train | | | | Valid | | | | Test | | | |
|---|---|---|---|---|---|---|---|---|---|---|---|---|
|  | B | I | E | S | B | I | E | S | B | I | E | S |
| LOC | 9395 | 4015 | 9395 | 991 | 1182 | 503 | 1182 | 124 | 1151 | 517 | 1151 | 155 |
| DATE | 599 | 388 | 599 | 699 | 66 | 38 | 66 | 88 | 72 | 40 | 72 | 75 |
| NUM | 151 | 32 | 151 | 3882 | 15 | 0 | 15 | 475 | 20 | 0 | 20 | 456 |
| PER | 281 | 16 | 281 | 1911 | 34 | 0 | 34 | 223 | 27 | 0 | 27 | 231 |
| ORG | 308 | 208 | 308 | 184 | 48 | 39 | 48 | 21 | 44 | 40 | 44 | 22 |
| TIME | 143 | 92 | 143 | 118 | 9 | 0 | 9 | 0 | 13 | 0 | 13 | 0 |
| O | 167547 | - | - | - | 21324 | - | - | - | 20967 | - | - | - |

### B. Conditional Random Fields

Conditional Random Fields (CRFs) are a class of probabilistic graphical models widely used for sequence labeling tasks, such as Named Entity Recognition (NER) and Part-of-Speech (POS) tagging. Unlike generative models, CRFs directly model the dependencies between labels, making them more effective for tasks where context is crucial. The traditional CRF approach relies on handcrafted features to capture contextual information. In our study, we use pretrained fastText word embeddings as a feature for traditional CRF training.

### C. Bidirectional Long Short-Term Memory

Bidirectional Long Short-Term Memory (BiLSTM) is a type of recurrent neural network (RNN) that processes input sequences in both forward and backward directions. This allows the model to capture contextual information from past and future tokens simultaneously, making it highly effective for sequence labeling tasks. Each LSTM unit maintains a memory cell and uses gating mechanisms (input, forget, and output gates) to control the flow of information. By combining the outputs of the forward and backward LSTMs, BiLSTM generates a rich representation of each token in the sequence. In our study, we use BiLSTM as the backbone of our neural models for NER and joint POS+NER tasks.

### D. Inference Layer

The inference layer predicts the final sequence of labels based on the features extracted by the BiLSTM. We explore two types of inference layers: CRF and Softmax. The CRF layer models dependencies between consecutive labels by considering transition probabilities between labels and emission probabilities (output scores from the BiLSTM), making it particularly effective for tasks like NER where label transitions follow specific patterns (e.g., "B-LOC" cannot follow "I-PER"). In contrast, the Softmax layer independently predicts the label for each token by applying a softmax function to the BiLSTM output scores, ignoring dependencies between consecutive labels. While simpler and faster, this approach may lead to suboptimal predictions for tasks with strong label dependencies.

## IV. EXPERIMENTS

For our experiments, single NER and POS+NER joint models were trained for each approach. We trained traditional CRF with and without fastText Embeddings as a feature for both single and joint models. For the word embeddings layer to train deep learning models, we used random word embeddings and both pretrained fixed and fine-tuned fastText embeddings. BiLSTM is trained for both single and joint sequence tagging with inference layers - softmax and CRF as shown in Figure 2.

### A. System Configurations

The system configuration used to train deep learning approaches took advantage of the Kaggle[1] environment with an NVIDIA P100 GPU. The P100 is a high-performance GPU with 16GB of memory, making it well-suited for training complex machine learning models. Its substantial memory capacity allows for efficient handling of large datasets and deep learning architectures. However, the P100 has a power consumption of 250W, which reflects its high computational capability. GPUs, such as the P100 and the T4 available on Kaggle, are versatile processors originally designed for rendering graphics but have become indispensable for parallel processing tasks like training machine learning models, enabling faster computations compared to traditional CPUs.

---

[1] https://www.kaggle.com/

### B. Hyperparameters

The hyperparameters for the experiments were carefully chosen to optimize the performance of both traditional CRF and BiLSTM-based models. The configurations for each approach are detailed as follows:

*1) CRF models:* The CRF models were trained using CRFsuite[2], leveraging both handcrafted features and word embeddings. For the traditional CRF models, two approaches were experimented - with and without word embeddings. The pretrained fastText embeddings had a dimensionality of 300 and were trained on the Wikipedia Burmese data [5]. To enhance pattern recognition in sequence labeling tasks, a variety of handcrafted features were included. These features capture both word-level and character-level information. At the word level, the context is considered by incorporating the preceding and succeeding words, as well as whether the word is the first or last in the sentence. At the character level, prefixes and suffixes of varying lengths (e.g., the first one to three characters and the last one to three characters of a word) are used to capture morphological patterns. Additional features include indicators for the presence of hyphens, numeric characters, and the overall position of the word in the sentence. These features collectively aim to improve the models ability to recognize complex patterns and dependencies within sequences.

*2) BiLSTM Models:* The neural models were implemented using PyTorch[3]. For the BiLSTM-CRF models, three types of word embeddings were used in the experiments. First, random embeddings were initialized with a fixed dimensionality of 300. Second and third, pretrained fastText embeddings with a dimensionality of 300 were evaluated in two configurations: fixed and fine-tuned. In the fixed configuration, the weights of the embeddings were kept constant during training, while in the fine-tuned configuration, the weights were updated as part of the training process.

For training with random embeddings, we used the BiLSTM layer with 128 hidden units and a batch size of 32. For training with fastText embeddings, the BiLSTM layer consisted of 256 hidden units in each direction, enabling the model to capture both forward and backward contextual information. A batch size of 64 was used for experiments with fastText embeddings.

In all experiments, to avoid overfitting, a dropout rate of 0.5 was applied to the embedding and BiLSTM layers. The Adam optimizer was employed with a learning rate of 0.001 and default parameters, specifically $\beta_1 = 0.9$, $\beta_2 = 0.999$, and $\epsilon = 1 \times 10^{-8}$. Training was carried out for a maximum of 50 epochs, with early stopping applied based on validation loss.

*3) Inference layers:* Two types of inference layers were evaluated. The first was a CRF (Conditional Random Field) layer, which modeled label dependencies by considering both transition and emission probabilities. This inference layer consistently demonstrated better performance, as reflected in the results. The second was a Softmax layer, which independently predicted the label for each token without modeling dependencies between consecutive labels. Although simpler and faster, the Softmax layer was less effective for tasks requiring strong label dependencies, such as Named Entity Recognition (NER).

### C. Evaluation

We evaluated NER performance only for both single-task and POS+NER joint-task approaches. The evaluation of our NER models' performance in our experiments relies on the following standard metrics:

- *F1 Macro Average*: The F1 Macro Average metric calculates the harmonic mean of Precision and Recall independently for each class and then averages the results. It treats all classes equally, regardless of their size, and does not consider class weights. This makes it particularly useful for evaluating the model's performance across all classes, especially when there is an imbalance in class distribution.

- *F1 Weighted Average*: The F1 Weighted Average metric calculates the harmonic mean of Precision and Recall for each class, weighted by the number of true instances in each class. This metric accounts for class imbalances, ensuring that more frequent classes have a larger influence on the final score.

- *Accuracy*: Accuracy measures the proportion of correctly predicted tokens out of the total number of tokens in the dataset. It provides an overall assessment of the models correctness at the token level.

These metrics were selected to provide a comprehensive evaluation of the performance of the model, taking into account both the general effectiveness and the specificity of the class.

## V. Results and Discussion

The evaluation of Burmese Named Entity Recognition (NER) models in this research highlights the impact of various training strategies on model performance, as presented in Table IV.

Through rigorous experimentation, key metrics such as **NER Accuracy** (Acc.), **NER Weighted F1** (F1-Wt.), and **NER Macro F1** (F1-Macro.) reveal significant findings regarding the effectiveness of our proposed approaches. Among the models tested, the Conditional Random Fields (CRF) models consistently performed well, particularly when integrated with fastText features, demonstrating the value of incorporating rich embeddings.

---

[2]https://github.com/chokkan/crfsuite
[3]https://pytorch.org/

TABLE IV: Performance Comparison of Single NER and Joint POS+NER Models (Best results are **bolded**.)

| Model Name | Embeddings | Training Type | Acc. | F1-Wt. | F1-Macro. |
|---|---|---|---|---|---|
| **CRF** | w/o fastText features | Single (NER) | 0.9818 | 0.9812 | 0.7405 |
| | | Joint (POS+NER) | 0.9812 | 0.9807 | 0.7367 |
| | with fastText features | Single (NER) | 0.9818 | 0.9811 | **0.7429** |
| | | Joint (POS+NER) | 0.9810 | 0.9804 | 0.7345 |
| **BiLSTM-Softmax** | + Random Embeddings | Single (NER) | 0.9740 | 0.9725 | 0.6478 |
| | | Joint (POS+NER) | 0.9730 | 0.9714 | 0.6463 |
| | + frozen fastText | Single (NER) | 0.9737 | 0.9723 | 0.6578 |
| | | Joint (POS+NER) | 0.9753 | 0.9734 | 0.6489 |
| | + fine-tuned fastText | Single (NER) | 0.9783 | 0.9779 | 0.6502 |
| | | Joint (POS+NER) | 0.9780 | 0.9764 | 0.6743 |
| **BiLSTM-CRF** | + Random Embeddings | Single (NER) | 0.9740 | 0.9730 | 0.6784 |
| | | Joint (POS+NER) | 0.9742 | 0.9730 | 0.6907 |
| | + frozen fastText | Single (NER) | 0.9747 | 0.9729 | 0.6396 |
| | | Joint (POS+NER) | 0.9746 | 0.9737 | 0.6790 |
| | + fine-tuned fastText | Single (NER) | 0.9755 | 0.9753 | 0.7154 |
| | | Joint (POS+NER) | **0.9791** | **0.9776** | **0.7395** |

The **CRF + fastText (Single)** model achieved an accuracy of 0.9818, a weighted F1 score of 0.9811, and a macro F1 score of 0.7429, which was the highest among all models tested, highlighting its ability to capture complex label dependencies and accurately identify named entities.

The **BiLSTM-CRF + fine-tuned fastText (Joint)** model also demonstrated impressive results, with an accuracy of 0.9791, a weighted F1 score of 0.9776, and a macro F1 score of 0.7395. This model, which combines bidirectional LSTMs with CRF for sequence labeling, outperformed other BiLSTM-based models and solidified the importance of joint training with Part-of-Speech (POS) tagging in enhancing the model's understanding of syntactic context. The success of this model underscores the potential of joint training in integrating syntactic and semantic knowledge for more precise entity recognition.

In contrast, while the BiLSTM-Softmax models, which employed randomly initialized embeddings or frozen fastText embeddings, showed competitive performance, they did not match the performance of the CRF-based models. The highest-performing BiLSTM-Softmax model, which utilized fine-tuned fastText embeddings, achieved a macro F1 score of 0.6743 in the joint POS+NER setting. This gap in performance illustrates the superior ability of CRF-based models to capture label dependencies, which is particularly important in sequence labeling tasks like NER.

The inclusion of POS tagging as a joint task proved beneficial across different model architectures. For example, while the **CRF + fastText (Single)** model outperformed the joint version in terms of macro F1 score, the joint training still maintained strong performance with only a slight reduction in accuracy and weighted F1 scores. On the other hand, joint training with BiLSTM models provided a more noticeable improvement in performance, particularly when fine-tuned embeddings were used. This further validates the importance of incorporating syntactic (POS Tag) information to improve NER performance, especially for languages like Burmese, which exhibit complex grammatical structures.

TABLE V: Tagwise F1-score predicted by fine-tuned fastText + BiLSTM-CRF (Joint)

| Tags | B | I | E | S |
|---|---|---|---|---|
| LOC | 0.9763 | 0.9711 | 0.9766 | 0.6849 |
| DATE | 0.8281 | 0.7654 | 0.8682 | 0.9136 |
| NUM | 0.4211 | - | 0.4211 | 0.9501 |
| PER | 0.9143 | - | 0.8986 | 0.5783 |
| ORG | 0.6265 | 0.5217 | 0.5977 | 0.5882 |
| TIME | 0.8889 | - | 0.8889 | - |
| O | 0.9897 | - | - | - |

Table V presents the tag-wise F1-scores predicted by the **BiLSTM-CRF + fine-tuned fastText (Joint)** model. The symbol "-" indicates cases where no corresponding tags were present in the test dataset.

Detailed tag counts can be found in Table II. Lower F1-scores for certain tags, such as B-NUM and E-NUM, can be attributed to the limited amount of data available during both training and testing. For instance, tags like NUM, which have relatively fewer occurrences in the training set (151 occurrences in B-NUM and 151 in E-NUM), contribute to weaker model performance due to insufficient data for the model to learn accurate patterns. On the other hand, tags with larger amounts of data, such as LOC (with 9395 occurrences in B-LOC and 9395 in E-LOC), tend to have higher F1-scores, indicating that more training examples help the model better generalize and improve its accuracy. This trend is consistent across the datasets, as shown in Table II, where higher tag counts in the training and test datasets correlate with better F1-scores.

The comparison between CRF and Softmax layers highlights the significant advantage of CRF in modeling label dependencies. The CRF-based models, both with and without fastText features, outperformed the Softmax models across all metrics. This finding emphasizes the importance of sophisticated inference mechanisms like CRF for achieving high-precision NER results, especially in morphologically rich languages where sequence dependencies play a crucial role. The research demonstrates that the **CRF + fastText (Single)** and **BiLSTM-CRF + fine-tuned fastText (Joint)** models represent the best solutions for word-level Burmese NER. These models achieved near-perfect accuracies of 98%, with **NER Weighted F1 scores of 0.98** and **NER Macro F1 scores of 0.74**, confirming the robustness and efficacy of our approaches in recognizing named entities with high precision, even within the constraints of a low-resource language setting. Our findings underscore the critical role of advanced training strategies, including the use of CRF inference layers and joint POS+NER training, in pushing the boundaries of NER performance.

## VI. Limitations

While the models presented in this research achieve high performance in Burmese Named Entity Recognition (NER), there are several limitations that should be considered. First, the model evaluation is based on a relatively small dataset, which may not fully represent the wide variety of linguistic structures and domain-specific entities found in the Burmese language. Although the incorporation of fastText embeddings and joint POS+NER training improves model generalization, the models may still struggle with handling low-frequency tags such as "TIME," "ORG," and "PER," and OOV words, particularly in specialized domains. Additionally, the class imbalance in the dataset, particularly for the "O" tag, could influence model performance and might require further techniques such as over-sampling or more advanced loss functions. Finally, while the CRF inference mechanism has shown superiority over Softmax, the computational cost and complexity of training CRF-based models may limit their scalability and deployment in real-world applications with very large datasets. For example, in our experiment, using our system configuration mentioned in Section IV, fine-tuned fastText + BiLSTM model with CRF inference layer cost 257.90 seconds but it only cost 34.58 seconds with softmax inference layer for training joint models.

## VII. Conclusion and Future Work

This research makes an important contribution to the development of Burmese Named Entity Recognition (NER) by evaluating CRF and BiLSTM-CRF models with different embedding strategies and training methods. The results clearly show that both CRF and BiLSTM-CRF models are effective at capturing contextual and sequential dependencies, making them excellent choices for NER tasks. A key finding of this research is the effectiveness of fine-tuned fastText embeddings, which improve model performance by capturing word-level details and helping with OOV words, especially for languages like Burmese that are complex and low-resource.

Additionally, this study confirms that joint training with POS tagging enhances the models' ability to understand language and improves NER performance without reducing accuracy. By combining both syntactic and semantic information, joint models demonstrate the power of multitask learning to improve NER, even in low-resource settings.

Another important result is the advantage of using CRF layers instead of Softmax layers. CRF layers are better at modeling the relationships and transitions between labels, which is crucial for tasks like NER. By combining these advanced techniquesCRF inference, fine-tuned fastText embeddings, and joint trainingthis research sets a new standard for NER performance in Burmese.

The findings of this research have broad implications, not just for Burmese but also for other low-resource languages. Future work can focus on integrating more linguistic features, exploring more advanced neural network architectures, and expanding datasets to cover a wider range of domains. These directions could further improve NER systems and make them more useful in other low-resource languages.